# Syntax and Semantics of Italian Poetry in the First Half of the 20th Century

*Rodolfo Delmonte* `delmont@unive.it`
*Dipartimento Studi Linguistici e Culturali Comparati, Università Ca Foscari, 30121 Venezia*

**Keywords:** `Italian Poetry 900;Rhetorical Figures;Non-canonical Syntactic Structures;Linguistic Theory;LFG;Computational Linguistics;Parsing Complexity;` **`Getarun system`**

## Abstract

In this paper we study, analyse and comment rhetorical figures present in some of most interesting poetry of the first half of the 20$^{th}$ century. These figures are at first traced back to some famous poet of the past and then compared to classical Latin prose. Linguistic theory is then called in to show how they can be represented in syntactic structures and classified as noncanonical structures, by positioning discontinuous or displaced linguistic elements in SpecXP projections at various levels of constituency. Then we introduce LFG – Lexical Functional Grammar as the theory that allows us to connect syntactic noncanonical structures with informational structure and psycholinguistic theories for complexity evaluation. We end up with two computational linguistics experiments and then evaluate the results. The first one uses best online parsers of Italian to parse poetic structures; the second one uses *Getarun* – the system created at Ca Foscari Computational Linguistics Laboratory. As will be shown, the first approach is unable to cope with these structures due to the use of only statistical probabilistic information. On the contrary, the second one – being a symbolic rule-based system - is by far superior and allows also to complete both semantic an pragmatic analysis.

## 1. Introduction

In this paper we will focus on Italian poetry of the first half of last century which is particularly attractive from a linguistic and rhetoric point of view. In Italy, the beginning of the 20th century saw a great number of movements in line with cultural development at European level in all fields: from Hermetism to Futurism. In one word what constitutes Modernism (through Crepuscolari, Vociani, etc.), which is deeply attracted by the linguistic experimentation of the Avanguarde. The poet had to surprise the reader, by the so-called Otstranenie mechanism proposed by Russian formalists, such as Viktor Shklovsky, Roman Jakobson and others.

Thus, a great number of Italian poets move away from traditional literary schemes. This applied both to the use of poetic devices and to the choice of the lexicon to increase linguistic expressivity associated to words. However, there is another dimension we are mostly interested in and it is the way in which the structures convey the meaning of the poem. We will concentrate on syntactic structure which is the vehicle both of meaning through semantics and of expressivity and emotions by the role of informational structure.

Topic and focus are activated by the pre/postposing mechanism of structural elements, thus generating unusual linear sequences of linguistic items which are hard to recompose. A borderline syntax of Italian has been used to focus on specific meanings and moods in order to produce (emotional) poetic effects in the poem. This is the theme of the paper which, starting from the syntax of the Italian poetry of first half of last century, highlights the use of non-canonical structures to obtain particular semantic and emotional effects.

Some such structures are for instance, Object Preposing and Subject Inversion, Clitic Left Dislocation and Subject Right Dislocation. These structures will be discussed from the point of view of linguistic theory in general and in particular with respect to the role they have in passing from form to meaning. The process of structure building will be looked at at first from a theoretical perspective, using LFG – Lexical-Functional Grammar, a theory that bridges the gap from abstract representations to psychological and computational processing.

In the final sections, the syntactic structures of some of the poems[1] will then be compared to the non-canonical syntactic structures present in the treebank of Italian called VIT (Venice Italian Treebank) that contains some 300,000 words

---

[1] All poetry examples will be taken from the Anthology of Poetry [20] published by Einaudi in 1994 and republished by L'Espresso

collected from texts ranging from 1980 to 2000 of contemporary written and spoken Italian(see [7]; [9]; [13]; [14]) . The comparison will allow to distinguish uncommon non-canonical syntactic structures in poetry from common ones, that can also be found in texts taken from daily newspapers and political commentaries, in literary prose and in other non-poetic genres.

We will then present computational linguistic experiments, where we will "parse" the poetic structures and the non poetic ones. From the analysis of four parsers of Italian we will try to ascertain whether noncanonical structures – both unusual ones from poetry and usual ones from common written text - are possibly parsed and if not why. To that purpose we will make a detailed analysis and comparison of the output of the best four online parsers in order to check what has been parsed correctly and the type of mistake they make. We will then end up with a section on *Getarun*, the system created at the Computational Linguistics Laboratory of Ca Foscari University that is organized in a pipeline of modules containing a parser but also semantic and pragmatic modules to mimick language understanding. *Getarun* also computes language complexity on the basis of computing or execution time. We will show some of the output from *Getarun* and explain its technique.

## 2. Descriptive vs Theoretical and Computational Linguistics

We will start our analysis from what are usually regarded rhetorical figures in poetry and part of descriptive linguistic approaches. These approaches describe phenomena characterized by special linguistic usage such as the ones listed below and then highlight the linguistic components without however explaining the structural processes underlying its deep nature. Typical descriptive labels for such rhetorical figures are the following ones: Enallage and Hypallage which are specifically meaning displacement at word level, through a transfer in the use of grammatical categories. And Hyperbaton (Tmesis) and Anastrophe (see [17:180]; [18:227]), which do the same but at a syntactic structural level. In our interpretation of these structures we will base our hypothesis on the book on Latin Word Order published by Devine, A. M. & Laurence D. Stephens, in 2006 [15] (hence D&L).

What we will then assume is that Italian poetry has inherited directly from Latin the possibility to create rhetorical figures which directly resemble Latin ones. In Italian, the first poet to use some of them consistently both at a linguistic and at a poetic level was Dante, followed by Petrarca and the other poets of the same period. In the small dataset we will be using, taken from more recent poets, there will be many examples that can only be traced back to Latin.

As D&L also assume, these rhetorical figures induce a displacement of information and are thus responsible for the semantic interpretation to assign to the stanza and the poem containing these non-linear or non-canonical structures. This interpretation hinges upon a syntactic theory and is strictly linked to semantics and pragmatics. From a theoretical point of view, syntax is regarded as the necessary intermediate level that allows to go from single words and sentences to their meaning. In our poetry, there is the need to redistribute semantic information in such a way to highlight or focus on some linguistic element in isolation rather than in their canonical position.

We will start now by presenting short excerpts made by just a few lines taken from the poems we intend to analyze. Some of the examples below are cases of hyperbaton, i.e. focussing by dislocation of portion of a constituent, typically an adjective from a noun phrase or a main verb from a verbal complex:

i. Adjective Extraction

"... Insieme sgraneremo altro rosario,

diverse pregheremo orazïoni,"

F.M.Martini, Vigilia di partenza, p.88

ii."... mentre ripenso, pallida, una gota,

mentre rivedo, piccola, un'effigie"

M.Moretti, La malinconia, p.94

iii. Lexical Verb Extraction + Subject Right Dislocation

"...E il tuo soggiorno un verde

giardino io penso, ove con te riprendere

---

in 2004.

può a conversare l'anima fanciulla,"

U.Saba, Preghiera alla madre, p. 336

In i. a quantified adjective is preposed out of the VP and of the Object NP orazïoni; in ii. the adjective is locally anteposed out of the Object NP and repeated twice; in iii a peculiar case which involves the main Verb riprendere which is preposed from the VP up in CP. Now hyperbaton is usually defined as a case of rhetorical figure in which two words which should stay close together are being separated by inserting other linguistic material. This definition is clearly defective of any linguistic information. Discontinuous or displaced constituents can be characterised by an extended number of different syntactic and semantic types which sometimes go against the principles of any linguistic theory. For example, Adjective extraction (anteposition) - which can be regarded as the best example of hyperbaton also in Latin - is not allowed in any linguistic theory. This principle was stated already in the '60s. It is formulated by saying that no constituent on the Left Branch of a Noun Phrase can be extracted from that NP. In particular, the adjective, belonging to the specifier of NP should stay there[2]. The example in iii. is even more complex to explain in theoretical term: a member of the verbal complex, the main verbal infinitival, presumably positioned inside a VP is extracted and anteposed, and still receives an interpretation.

We will show below appropriate linguistic descriptions for all of these cases. What is important to note, is the suspension of the meaning creation process that intervenes every time one of such examples is encountered. The reader is obliged to stop and reconsider the previous lines before continuing reading. We show now other examples taken this time from classical Italian literature which can be regarded are as extreme as the ones we saw before.

iv. io parlo de' begli occhi e del bel volto,

che gli hanno il cor di mezzo il petto tolto.

(L. Ariosto, Orlando furioso, C. VIII, vv. 639-640)

v. i ritrosi pareri e le non pronte

e in mezzo a l'eseguire opere impedite.

(T. Tasso, Gerusalemme liberata, C. I, vv. 235-236)

vi. il divino del pian silenzio verde.

(G. Carducci, Il bove, v. 14)

vii. Quelle sere, Maria non, come suole,

pregava al mio guanciale, co' suoi lenti

bisbigli, con le sue dolci parole:

(G. Pascoli, La mia malattia, vv. 11-13)

In iv. the auxiliary hanno has been separated from its main verb tolto by inserting the Object NP and the Adverbial SP. In v. the negated adjective non pronte is separated from the head noun it modifies opere by inserting the adverbial adjunct in mezzo a l'eseguir. In addition, the predicative past participle impedite is dislocated at the end of the sentence and separated from the conjunction e. In vi. it is again an adjective divino which is separated from its head noun silenzio by the preposed SP modifier del pian. Eventually, in vii. the negation non is separated from its verb by a parenthetical come suole. All of these examples are non-canonical structures which cannot be found in current Italian (literary) prose nor in spoken conversation. Separating auxiliary from participle main verb is allowed in ordinary prose only if adjuncts are inserted: complements are not allowed, Italian not belonging to the class of V2 languages like German[3]. The same applies to anteposed adjectives and negation.

Here below we list cases of Anastrophe that involve complete constituents. Some of these cases that can also be found in standard non-poetic texts.

viii. Subject Right Dislocation

---

[2] The Left Branch Condition was formulated by Ross ([19:127]) proposed the Left Branch Condition (LBC), which blocks movement of the leftmost constituent of an NP. The condition has been used in the literature to block extraction of determiners, possessors, and adjectives out of NP. However, Ross noted that the LBC is allowed in Latin and some other language.

[3] As for instance in "Wir haben dein Buch gelesen", literally "we have your book read"and reordered in English "we have read your book".

...giunse col vento un ritmo di campana,

G.Gozzano, La signorina Felicita (ovvero la Felicità), p. 47

ix. Object Right Dislocation

"...poi al vespero richiamo battendo il tamburello

sulla piazzetta il popol rusticano,"

G.P.Lucini, La parata dell'introduzione, p.114

x. Object Preposing

"...se, primavera, il mio cuor generoso

soffocasti di spasimi sordi"

C.Rebora, da Frammenti Lirici - XXI, p.203

In viii. the Subject NP un ritmo di campana is extraposed at the end of the sentence; in example ix. it is Object NP il popol rusticano to be dislocated or scrambled after the Adjunct gerundive; finally in x. we have a case of Object preposing, but in an intermediate position after Subject NP primavera, which could be regarded an Hanging Topic. All of these three cases can be found in standard Italian prose both in newspaper and in some specialized genre, like for instance bureaucratic documents or literary writing, as will be shown in a final section of the paper.

Finally, here are the last three examples we want to analyze, which like hyperbaton's cases above, can only be found in poetry:

xi. Negated Object Preposing+SP Modifier Dislocation+Subject

"...né i melliflui abbandoni

né l'oblïoso incanto

dell'ora il ferreo bàttito concede."

C.Rebora, da Frammenti Lirici - XXI, p.203

xii. Preposed Object-Adverb-Subject-Verb Structures

"... Lei sola, forse, il freddo sognatore

educherebbe al tenero prodigio"

G.Gozzano, La signorina Felicita (ovvero la Felicità), p. 49

xiii. Predicative Adjective Postposition + PP Modifier stranding

un'eco di mature angosce

rinverdiva a toccar segni

alla carne oscuri di gioia.

(S. Quasimodo, S'udivano stagioni aeree passare, vv 3-5)

In xi. we see a repeated conjoined and negated double Object NP fronting, followed by another discontinuity, dell'ora a PP subject modifier of the Subject NP il ferreo battito, and eventually the main verb concede; in xii. the preposed Object NP Lei sola where the pronoun Lei postmodified by a focal adjective is followed by an Adverb, forse and then the regular IP structure; finally in xiii. we have a Complement of an Adjective alla carne which is preposed, but also a PP modifier di gioia which is extraposed from its head NP headed by segni, out of the Infinitival clause. In the following section we will describe our theoretical approach both at a linguistic and at a computational processing level.

# 3. From Syntactic Structure to Semantic and Pragmatic Interpretation

Parsers[4] of Italian would be trained (statistical ones) on linguistically annotated collections of texts or corpora; or would contain rules (symbolic ones) making up a grammar for standard Italian structures, which may include cases of discontinuity, dislocation and fronting of constituents. But they will be unable to cope with the cases of hyperbaton and anastrophe shown above. This is because they will reject cases of fronted adjectives which need to be interpreted as local modifiers of their head noun: the same would apply for main verb extraction. As a result, these structures will prevent semantic compositional interpretation from taking place and the parsing process will stop and fail. At the same time, we might assume that these exceptional structures will trigger non compositional interpretations and no reconstruction is needed. However, since the poet has intentionally placed the adjective/verb in that position to create rhetorical effects, dismissing the discontinuity as a mere syntactic fact would be a mistake.

In other words, following D&L, we will treat pragmatic non-canonical and non-standard structural cases as compositional. They will all be treated as encoding pragmatic (informational) meaning that has to be reconstructed. To that aim, we will start by assuming that those unusual structures are allowed in Italian – and only in Italian - because of its strong Latin inheritance. In fact, as will be shown below, we will be able to find almost identical structures in Latin text or poem, which are reported and commented in D&L. We will thus follow their suggestion to characterize each structure at first using a standard X-bar syntactic constituency-based representation. And then, we will comment its semantic and pragmatic content also using parallel Latin structures when available.

First of all we will assume that Italian is a (weakly) configurational language with a neutral or canonical sentence structure of type SVO (Subject-Verb-Object). In fact Italian can also be regarded as a weak non-configurational language thanks to the presence of some important typological parameters: free subject inversion, pro-drop and non-lexical expletives(see [6])[5]. Since we follow LFG – Lexical-Functional Grammar – as our syntactic theory, the presence of non-canonical structures will not be presented using the movement metaphor. Constituents or subconstituents in non-canonical positions are assigned one of two discourse functions, Topic and Focus and will be accordingly positioned under CP, under IP, or under VP. Differently from what happens in a Chomskian framework, syntactic positions in LFG do not have any intrinsic relation to meaning interpretation. As a consequence, non-canonical structures – be they sentence internal or sentence external – are all interpreted locally. In LFG functional projections may host non-argument or discourse related constituents and consequently they can host TopicP and FocusP. They will also host operators and other non-argument linguistic material like complementizers and subordinating conjunctions. On the contrary, argumental projections are all lexically governed. In addition, LFG makes a neat distinction between predicative or open grammatical functions vs closed ones. The former allow some form of control from the outside, as happens for instance in infinitivals; the latter are semantically complete structures.

In LFG there are two levels of syntactic representation which consist of c-structures and sets of attribute-value pairs called f-structures. F-structures are built up from the syntactic surface structure of a sentence by unification of the attribute-value graphs that are associated with each word and each constituent of the c-structure of that sentence. LFG has a constraint-based, parallel correspondence architecture; it has no serial derivations (unlike transformational grammar); there are no deep structures. Abstract relations are locally distributed as partial information across words and overt fragments of structure, and may be monotonically synthesized in any order or in parallel. In that sense, interpretation in LFG may be regarded as noncompositional. In approaches which make use of surface structure interpretation of phrase structure trees, the order of the arguments is determined in part by the linear order of constituents, and in part by the hierarchical structure of constituents. F-structures have dominance but not precedence defined on them  – more on this below. The information in phrase structure trees that is represented in terms of linear precedence is encoded by the attribute names of the f-structures where for instance, the OBJect function precedes the IndirectOBJect and the OBLique. C-structure may vary from one language to another and is based on X-bar grammar[6].

---

[4] A parser is a highly sophisticated computer program that is used to produce a linguistic representation of a sentence or a text. The representation usually obeys some linguistic theory and can be used for practical applications like Question/Answering. The most common linguistic theory today is Dependency Grammar Theory. State of the art parsers are currently trained on converted versions of Penn Treebank into dependency representations, which however don't include null elements. This is done to facilitate structural learning and prevents the probabilistic engine to postulate the existence of deprecated null elements everywhere. However it is a fact that in this way the semantics of the representation used and produced is inconsistent and will reduce dramatically its usefulness in real life applications, like Q/A and other semantically driven fields, by hampering the mapping of a complete logical form. What systems have come up with are "quasi"-logical forms or partial logical forms mapped directly from the surface representation in dependency structure.

[5] It also has no wh- in situ, no preposition stranding, no deletable complementizers, no impersonal passives, no parasitic gaps with the same argument.

[6] In fact we will adopt what is currently called "Complement Syntax" which is so defined by D&L, p.87: "We call this type of syntax "Complement Syntax". The noun phrases appear after the verb in the neutral order DO-IO-Obl-Adjnct, and that is the order in which they compose with the verb. In the traditional structure for complement syntax the further to the right the argument or adjunct in the serial order, the higher it is in the tree. This accords with the compositional order for arguments relative to each other (modulo disagreement about the thematic hierarchy), for arguments relative to adjuncts, and with the scopal properties of adjuncts relative to one another. Arguably, the general principle is that information is built up in order of increasing restrictiveness and specificity (in terms of the speaker's default conceptualization of events). However, this structure creates problems for those theories in which

For ease of representation, we will adopt the following grammar which is reminiscent of the pre-minimalist Chomskian approach:

CP → SpecCP, C'
SpecCP → TopicP , FocusP
C' → C0, IP
C0 → Complementizer
IP → SpecIP, I'
SpecIP → SubjNP
I' → I0, VP
I0 → InflVerb (Neg, Aux, Mod, Asp)
VP → SpecVP, V'
V' → V0, ComplVP
V0 → (PartV)
ComplVP → ObjNP, OblPP, AdvP, CP

**Figure 1.** Standard rules of an X-bar grammar

Where we see properties of X-bar theory which are easily recognizable and are: Specifiers are always sisters to X' and daughter of XP. I' projects IP, I projects I'. Complements are sisters to X and daughter of X'; and Adjuncts are sister of X' and daughter of X', as it is also shown in Figure 2 below (see [5]). C-structure representation will make use of this X-bar syntax which may differ from one language to another.

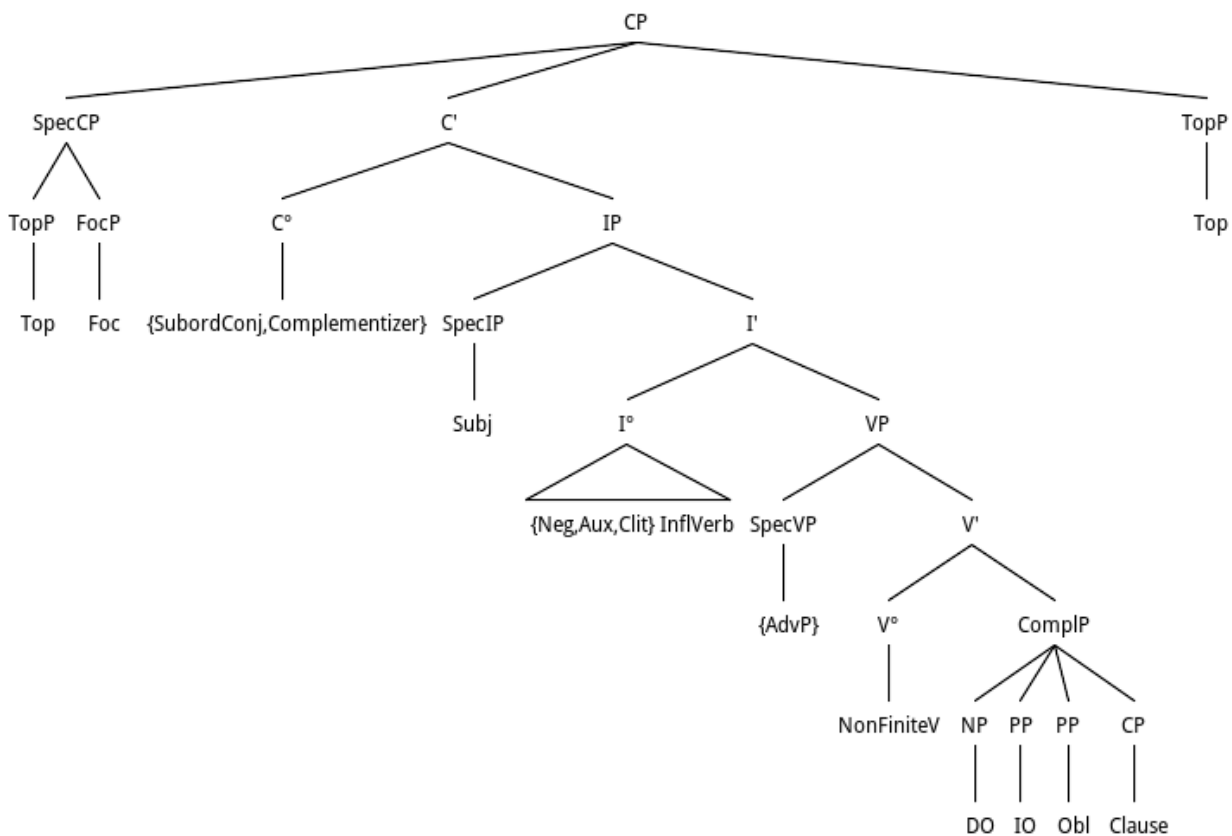

**Figure 2.** Tree representation of Topic Focus according to X-Bar theory and Complement syntax

binding and negative polarity are taken to be sensitive to c-command in the syntax. To deal with this latter problem, many linguists currently posit a different structure (instead of or in addition to the traditional right-higher structure), in which the further to the right an argument or adjunct is, the lower is in the tree. To achieve this, all phrases except the last one are assigned to a specifier position rather than a complement position, and each specifier is associated with a null-verbal head in a structure known as a VP shell."

In Italian, TopP and FocP can also appear under IP and under VP, and Subject NP can be postponed inside the VP or extraposed outside the sentence to the right – as an afterthought. C-structure heads (tree-node) correspond to f-structure (feature structure) heads. Also mapping regularities may apply: specifiers may be filled by discourse related functions SUBJ, FOCUS, TOPIC. Here I am using a configurational representation of Topic and Focus to show in a simpler manner the possible positions these non-argument functions may take. However, as said above, the syntactic position has no consequence with respect to its interpretation. In fact, it will be just the lack of an argument constituent that will trigger its interpretation as discourse function and the insertion of a variable or presence of a pronoun in its place inside the sentence. A Focus/Topic will thus bind an argument empty function (a Focus typically an OBJ; a Topic typically a SUBJ) and add its pragmatic content to it – more on this in the section below. The representation of discourse functions FOCUS and TOPIC in annotated c-structure looks like this:

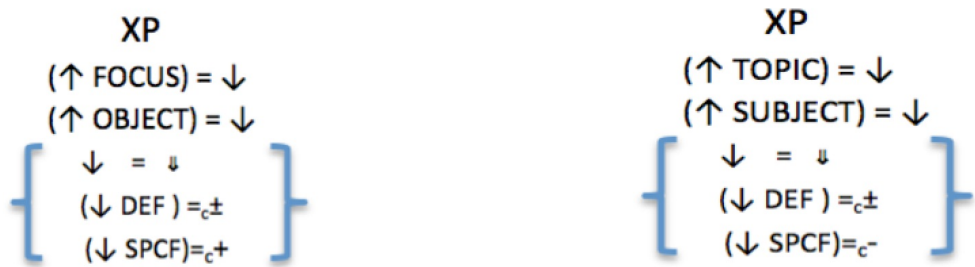

**Figure 3.** C-structure Annotation for Focus and Topic non-argument relation

where the equations are used to indicate the local content of the XP constituent, but also the semantic constrains that apply to it, i.e. that it must be specific and indefinite or definite, i.e. the determiner must be expressed unless it is a mass noun; the opposite applies to TOPIC functions. Current versions of LFG propose the following architecture of its internal projections, where the A-structure is the Argument structure projected to F-structure from the lexicon which imposes its contents by means of the Lexical Mapping Theory (see [3]; [4]). It is used to map semantic roles to grammatical functions according to a Thematic Hierarchy and a list of constraints. The lexicon contains subcategorization requirements which are encoded as semantic form with an argument list as shown below:

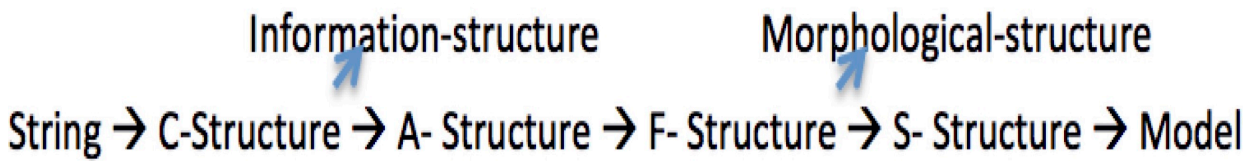

**Figure 4.** Modularized pipeline for string processing according to LFG theory

Information structure would contain a parallel structural representation for non-argument functions like Topic and Focus[7]. All the levels of grammatical representation in the projection architecture are simultaneously present due to unification. Mapping from c-structure to f-structure also imposes wellformedness conditions by foundational grammaticality principles of the LFG grammar that we report below:

*Uniqueness* or consistency, each f-structure attribute has at most one value. ([16:65]). In other words, each use of a semantic form is unique.

*Coherence* An f-structure is *locally coherent* if and only if all the governable grammatical functions that it contains are governed by a local predicate. An f-structure is *coherent* if and only if it and all its subsidiary f-structures are locally coherent.( [16:65]). In other words, no arguments which are not listed in the semantic form may be present.

*Completeness* An f-structure is *locally complete* if and only if it contains all the governable grammatical functions that its predicate governs. An f-structure is *complete* if and only if it and all its subsidiary f-structures are locally complete.( [16:65]). In other words, all arguments which are listed in the semantic form must be present.

---

[7] The architecture shown here is inspired by [1]. A-structure stands for argument-structure where predicate-argument structures are visualized with their semantic roles. S-structure stands for semantic-structure where semantic analysis has been carried out – for instance, pronominal binding, anaphora resolution and quantifier raising. Model is where entities, their properties and attributes and their relations are recorded with their semantic indices – more on this below.

In the following section we will represent rhetorical figures using c-structures and commenting their peculiar type of discontinuity, dislocation or extraposition also on a computational basis. Then in Section 5 we will try to motivate the use of LFG to connect form and meaning.

## 4. The Syntax of Hyperbaton and Anastrophe

Rhetorical figures found in poetry are totally different from and cannot be used in normal or even specialized or literary prose. So the idea to refer to Latin is due to the fact that some of these exceptional structures are very common in Latin prose as D&L discuss at length – and can also be found in other languages like Serbo-Croatian(see [2]). Consider a simple case from Caesar's De Bello Gallico, Book 5, paragraph 53, magnas Gallorum copias/a great quantity of Gauls/ reported at p.385 and reproduced here below:

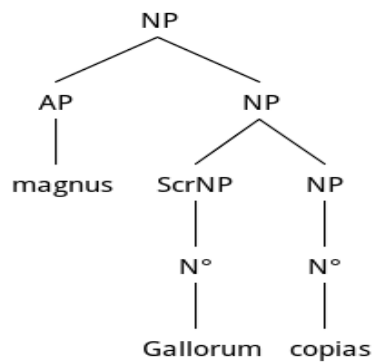

**Figure 5.** Tree structure of a Latin hyperbaton

This can be compared to the example reported above under vi. and repeated here, il divino del pian silenzio verde, where we have a case of Genitive Hyperbaton. We call the raised genitive Gallorum Scrambled (Scr) NP, following D&L. Again, there is no theoretical implication in the use of that term which I introduce again for clarity in the representation.

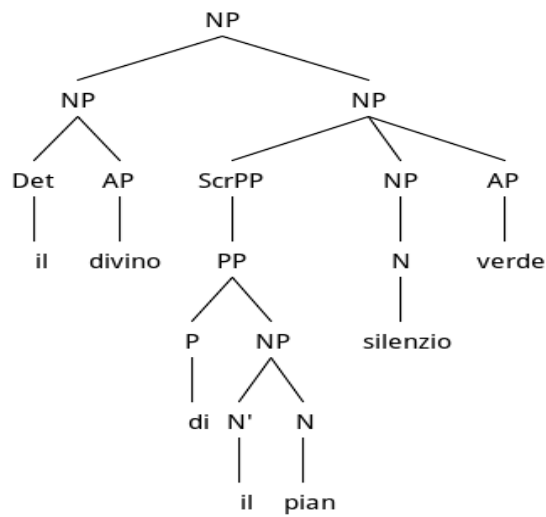

**Figure 6.** Hyperbaton and fronting of PP subject from Carducci

In Figure 6 we show the fronting of the Subject PP "del pian"/of the plane/ from the end of the NP in an intermediate NP position: this is not the Specifier position which is still occupied by the Determiner definite article "il". The problem with this representation is the separation of the head "silenzio"/silence/ from the determiner by presence of two additional projections: an AP and a PP. These two projections could both be positioned under a unique Specifier: on the contrary we chose to represent what could constitute an incremental analysis. When the adjective is reached, it could be taken as the current head of the NP: but then additional material arrives and a new Specifier position is represented by the anteposed PP "del pian"/of the plane/. However, note that the adjective has to be interpreted as a separate constituent in order to become a Focus.

Besides, the overall NP could have been expressed by at least the four following linear orders:

a. il divino silenzio verde del pian where we positioned green to modify silence rather than plane as a predicative modifier producing a synaesthesia effect;

b. il divino silenzio del verde pian where both adjectives are used in attribute position and modify their head nouns in an usual semantically natural way – as meaning restrictions, even though divino/divine/ is an abstract property which is not usually associated to silence;

c. il divino silenzio del pian verde less likely due to the apocope on the noun piano, and

d. il silenzio divino del verde pian which represents the more likely linear order, where however, divino is moved from attributive to predicative position.

From a semantic point of view, the question is how to reconstruct the focussed adjective with the noun it modifies, and this could only be done after the PP "of the plane" has been interpreted as the nominal Subject of the head noun "silence". In other words, interpretation of the adjective "divino" - which should be contiguous to the noun it modifies - has to be suspended and put on a Focus stack in order to be associated to its head. An adequately labeled representation could be the one under Fig.6a where fronting is defined as sequence of Focus and Topic.

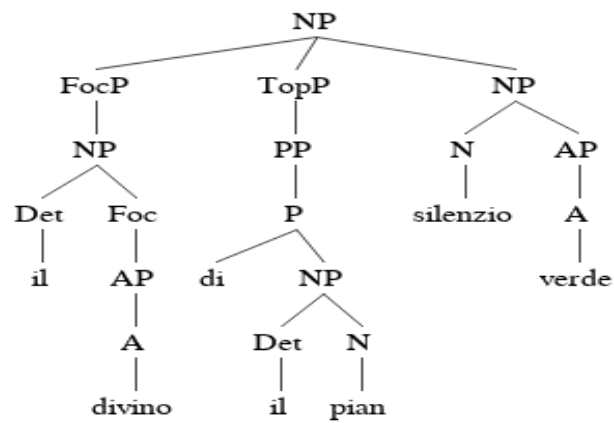

**Figure 6a.** Hyperbaton and fronting of AP and PP as Focus and Topic

Another similar case is represented by the fragment presented under i. and here under Fig.7. However, in this case the focussed adjective "diverse" can be represented under SpecCP due to the possibility to leave the Subject lexically unexpressed – but still understood as an empty pronoun with verb inherited morphology.

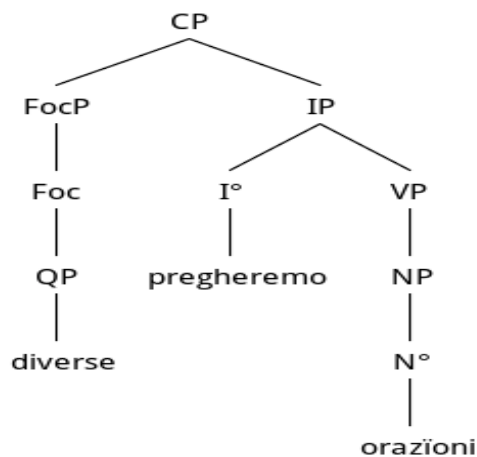

**Figure 7.** Hyperbaton and fronting of quantified adjective "diverse" from Object NP

Adjective extraction in this case is done from inside a VP complement, the Object NP orazïoni. In this as well as in cases reported in Figg.5 and 6, agreement is important to constrain restructuring. However, as said above, in both cases focussed material is interpreted locally as carrying additional pragmatic meaning to the one obtained by compositional operations. The reordered version is pregheremo diverse orazioni/we will pray several prayers/

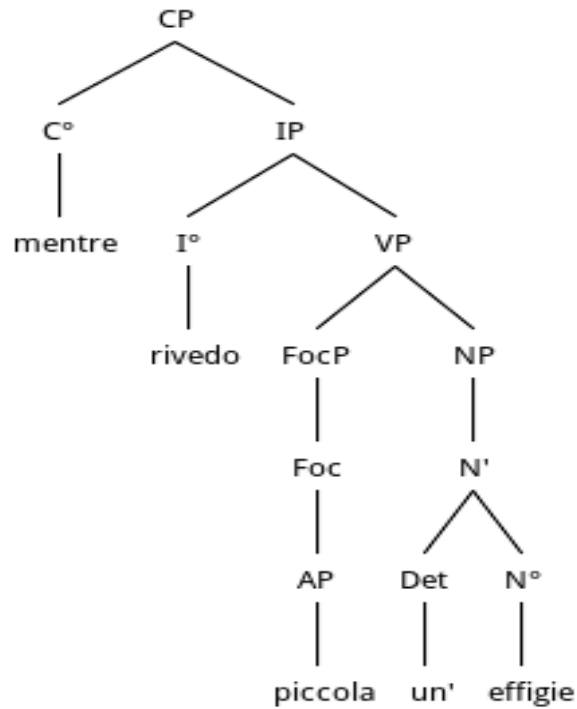

**Figure 8.** Hyperbaton and fronting of quantified adjective "piccola" from Object NP

In the example reported in Fig.8, Adjective extraction is operated locally in the Object NP, after the main governing verb has been expressed and will thus be represented in the specifier of VP. From a processing point of view, the presence of a determiner – the indefinite article una – imposes this structural choice. Note again that the adjective has to be interpreted as a separate constituent in order to become a Focus. This may be possible if we consider that the adjective might be Predicative and also that it may become an Adjunct. Here is the reordered version and its translation: mentre rivedo una piccola effigie/while I see again a small image/

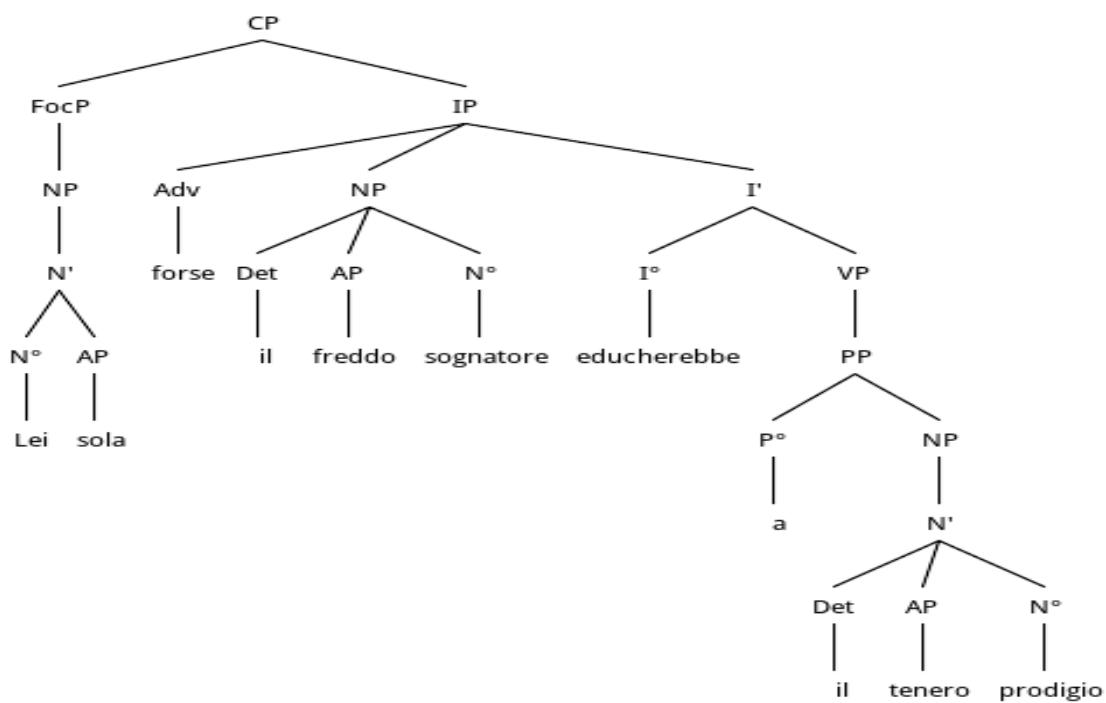

**Figure 9.** Anastrophe and fronting of Object NP from VP

In Fig.9 we represent a case of Object fronting which is reinforced by the insertion of the focal adjective sola and is prosodically separated from the rest of the clause by the presence of the adverb forse. Here is the linearly reordered version of the sentence: Il freddo sognatore forse educherebbe lei sola al tenero prodigio/The cold dreamer maybe would educate her alone to the tender prodigy/. From a computational point of view, this case requires the parser to fail in the attempt to process a canonical SVO structure. After failure, the parser will try a sequence of non-canonical structures, including the one above. From an informational point of view, FocP should come before TopP, which is represented by the Subject NP.

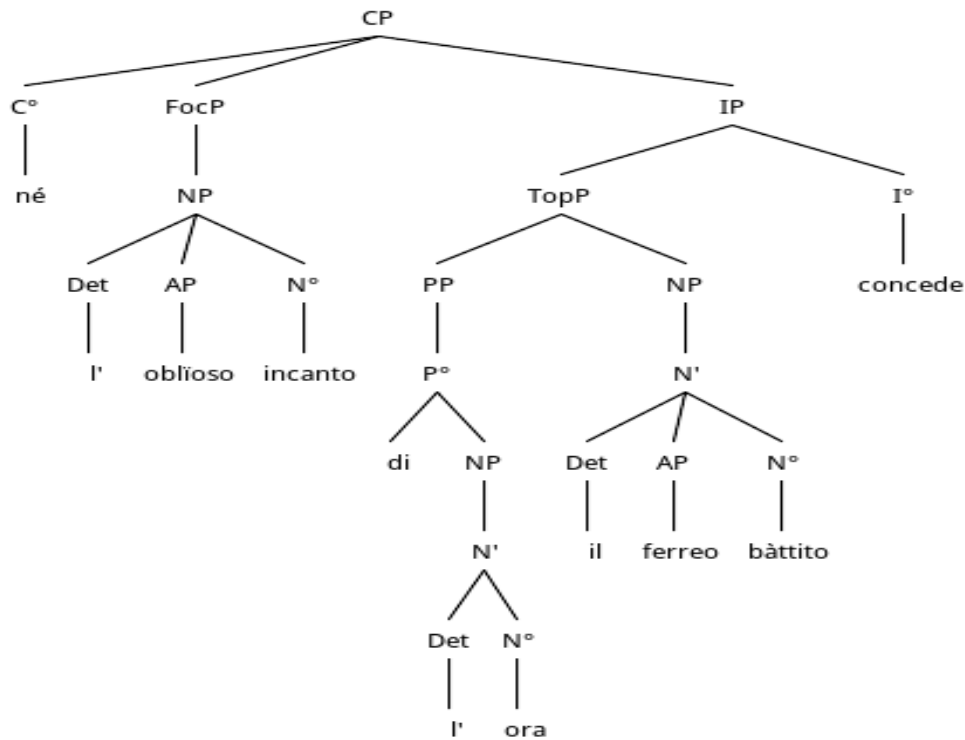

**Figure 10.** Anastrophe and fronting of Object NP from VP and PP Subject modifier from NP

In Fig.10 we have another example that D&L have defined as Genitive Hyperbaton as in animi incitatio/excitement of the mind/ (De Part Or 9) quoted in (1) p.525 or pacis umquam apud vos mentionem/mention of peace to you/ (Livy 21.13.3) quoted as (2) p.526. The example also contains a FocP as we saw in Fig.9. This is the reconstructed linear version of the sentence: né il ferreo battito concede l'oblioso incanto dell'ora/nor does the iron beat allow the oblivious enchantement of the clock-time/. We labeled the Subject "Il ferreo battito" as TopP and placed it under IP even though it contains the scrambled genitive modifier "dell'ora" – which is in fact to be interpreted as the nominal Object of the head "battito".

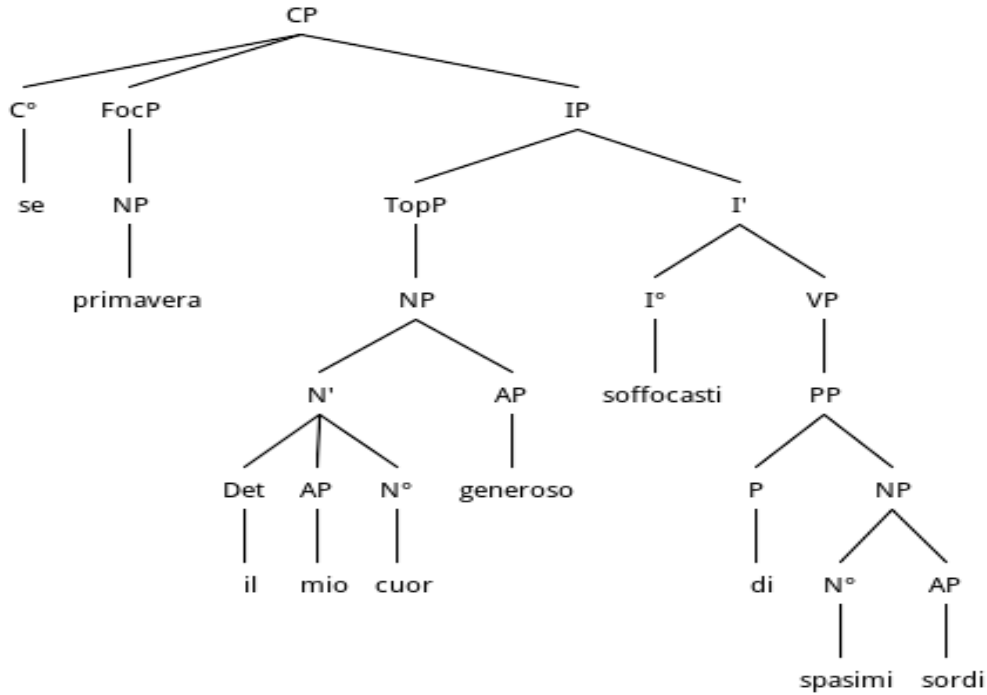

**Figure 11.** Anastrophe and fronting of Subject NP and Object NP from VP

In Fig.11 we have a NP "primavera" - a vocative apposition reinforcing the unexpressed subject "tu" - that has been positioned in front of the Object NP, preposed in front of its governing verb "soffocasti". The higher NP can thus be interpreted as a case of Hanging Topic coreferring to the unexpressed subject pronoun, also due to the presence of commas around it. Here the complete linear sentence is as follows: "se, primavera, (tu) soffocasti di spasimi sordi il mio cuor generoso"/if, spring, you choked with hollow spasms my generous heart/. It is important to remark the lack of agreement between the vocative higher NP "primavera", 3rd person singular, and the unexpressed pronoun subject of the governing verb "soffocasti", "tu", 2nd person singular. A better linear version of the sentence would thus be this one, "se (tu), primavera, soffocasti di spasimi sordi il mio cuor generoso", where the apposition is placed correctly after the unexpressed subject pronoun "tu".

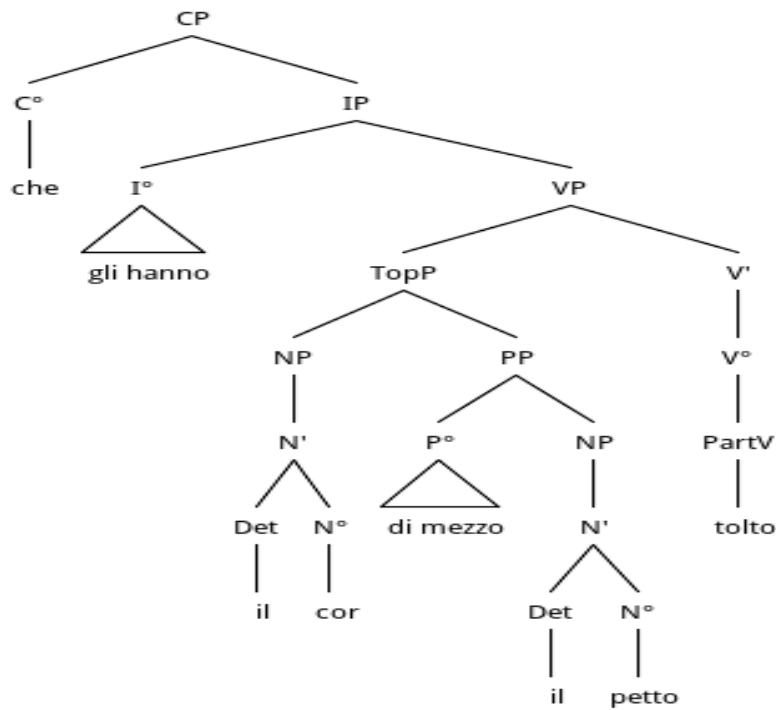

**Figure 12.** Anastrophe and Hyperbaton fronting of Object NP and Oblique PP from VP – the auxiliary is separated from past participle main verb

The example in Fig.12 is a case which could be described as Hyperbaton and as Anastrophe at the same time. The splitting of Auxiliary and main verb is a case of hyperbaton, but the anteposition of object and oblique are both cases of anastrophe. They are all understood as Topicalization, that is anteposition of known information. The linear version of the sentence is as follows: che gli hanno tolto il cor di mezzo al petto/which have torn the heart from the middle of the chest/. From a computational point of view, this is only possible if the past participle is reconstructed at the end of the clause it belongs to. The parser has already computed the auxiliary hanno as main lexical verb. But when it completes the clause, the past participle tolto remains rejected by the rules. It is thus recovered before building the clause level c-structure represented and substituted to the auxiliary verb, which however, contains all the morphological features of Tense, Mood, Person and Number.

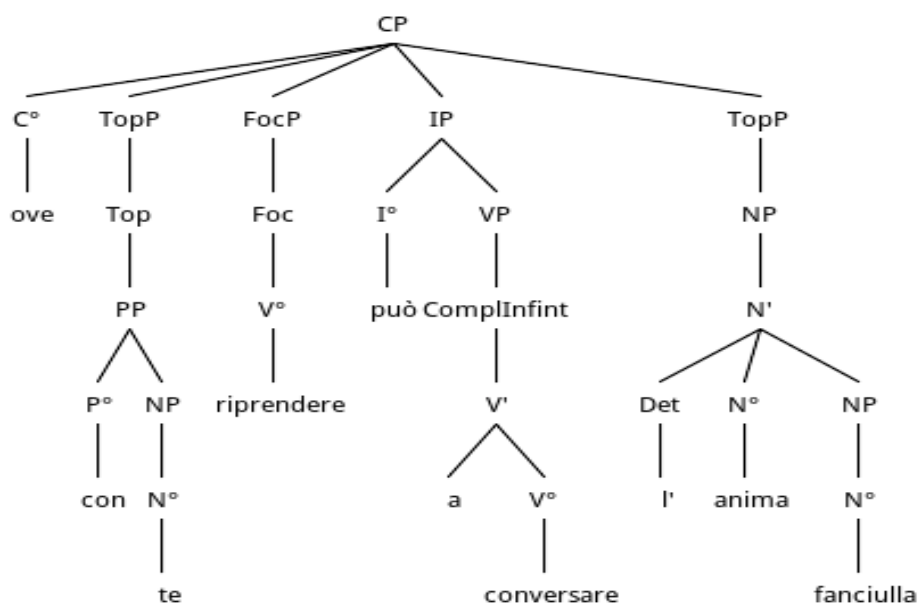

**Figure 13.** Anastrophe and Hyperbaton fronting of Comitative PP and Infinitive Main V from VP and complement VP, and Subject Extraposition

Eventually, in Fig.13 we have another complex example where there is fronting of a Comitative PP as Topic extracted from a VP Infinitival complement; then an extraction of a main nonfinite verb riprendere from VP and at the so-called right periphery we have the extraposed or dislocated Subject NP. The linear version of the sentence is the following one: "ove l'anima fanciulla può riprendere a conversare con te"/where the girl soul can restart to converse with you/. From a computational point of view, the main nonfinite verb is easily reconstructured inside the Verb Complex, being adjacent to the modal auxiliary, in reversed position. The reconstruction of the fronted PP inside the infinitival is more difficult. The right dislocated Subject NP is easily computed being part of a non-canonical structure which, as happened previously, is accessed after failure of the canonical sentence structure.

# 5. LFG and Discourse Non-argument Grammatical Functions

In our approach we have always indicated LFG as the theoretical and practical backbone of our research activity, which has inspired also heavily the way in which our computational work has been carried out[8]. This is due to the choice of LFG to support a psycholinguistic approach in which performance played an important role and to call for a processor as a fundamental component of the overall theory([16:xxii-xxiv]). LFG is a transformationless theory which is based on the lexicon and the existence of lexical rules to account for main NP movements. Long-distance dependencies are accounted for by properties of the f-structure. A displaced f-structure, receiving one of the non-argumental pragmatically related functions, FOCUS or TOPIC, is made dependent or fuse with the missing function in a following or preceding f-structure that requires it. Requirements are dictated by grammatical principles of Completeness and Uniqueness and the Extended Coherence Condition(ECC) by which FOCUS and TOPIC must be linked to the semantic predicate argument structure of the sentence in which they occur, either by functionally or by anaphorically binding an argument formulated by [1]. A possible filler or head is functionally equated with some argument in order to satisfy the ECC. Eventually the gap will be filled by the content of the non-argument discourse function which will thus occupy or consume two grammatical functions. However this operation will require a variable binding operation or an anaphoric control for resumptive pronouns, once the f-structure unification mechanism reaches the governing verb at clause level. Empty arguments will be filled by a variable or an empty pronoun, in case the language allows it – Italian does it mainly for SUBJect pronouns, which are called little-pro.

---

[8] The system Getarun was presented at one of LFG annual conferences in Athens in 2006 and is available here, https://pdfs.semanticscholar.org/ae25/5641cc68ec2fc9409abf754511551fa1784a.pdf

At a psycholinguistic level, this operation induces additional complexity. Let's see how the working of the mental processor can be described. To suit the limitations imposed by Short-Term or working Memory, no more than 7 single linguistic items can be stored before they enter Long-Term Memory. This is mimicked by the working of an incremental parser that computes fully interpreted structures which in LFG should correspond to F-structures. In our previous work([10], Chapter 3) we presented a principle-based version of the parser that takes advantage of the Minimalist Theory (hence MT) to instruct the "processor" while inputting words incrementally in the working memory. In the sections above, the theory we followed was LFG but in both cases it is now the lexicon that drives the computation: in fact, lexical information in the MT makes available features to the processor that will use Merge and Move to select appropriate items and build a complete structural representation. To justify memory restrictions, the parser takes advantage of an intermediate level of computation, called PHASE constituted by a fully realized argument structure preceded by a Verb. Reaching the verb is paramount also for LFG theory, for selection but also for grammatical function assignment. In [12] we reported experiments in which the moves of an MT-inspired processor are accompanied by reading times, which seem to (partially) confirm the prediction of the theory. In turn, these predictions are computed on the basis of local and non-local features and depend strictly on the type of argument selection operated by the verb.

In addition, we assume that referential properties of NPs as shown at determiner level may induce different cost measures: definite NPs being heavier to process than Proper Nouns, and these in turn heavier to process than deictic personal pronouns (you/I). This hierarchy is then partially reinforced by presence of distinct features. The underlying idea is that higher costs are related to the integration of new referential material which needs to be coreferential with previously mentioned antecedents: this is regarded heavier but on the same level of third person pronouns, followed by easier to integrate Proper Nouns that can be identified uniquely in the world. In the experiments reported in the section below, we will compute such measures in sentences containing the three types of referring expressions: definite NP, proper noun and personal pronoun.

Following this line of reasoning, we take complexity measures to be sensible to non-canonical structures that are pragmatically motivated and are used to encode structured meaning with high informational content, related to the FOCUS/TOPIC non-argument functions in LFG. Non-canonical structures can be said to help the reader or interlocutor to better grasp the intended (pragmatically) relevant meaning in the context of use (see [12]). Predictability of a certain noncanonical structure (hence NCS) may also depend on its frequency of use in given contexts. Italian noncanonical structures are relatively highly represented, as the following table shows, as extracted from VIT, our treebank [9], where they have been explicitly marked with the labels indicated below:

| NCS/Types | LDC | S_DIS | S_TOP | S_FOC | DiscMods | Total | % Non Project. | % NCS/ TSSen |
|---|---|---|---|---|---|---|---|---|
| Counts | 251 | 1037 | 2165 | 266 | 12,437 | 16,156 | 7% | 84.59% |

**Table 1.** Non-projective/noncanonical structures in VIT divided up by functional types

The final percentage is computed on the total number of constituents in written texts, amounting to 230,629[9]. If we compare these data with those made available for Latin (see [12]), where the same index amounts to 6.65% - data taken from the Latin Dependency Treebank containing some 55,000 tokens -, we can see that Italian and Latin are indeed very close. The second percentage is computed by dividing up number of NCS/TotalNumber of SimpleSentences. As for tree projectivity in the Penn Treebank (here marked as PT), numbers are fairly low as can be seen in the following table.

| TBs/NCS | NCS | UnxSubj | TUtt | TSSen |
|---|---|---|---|---|
| VIT Totals | 3,719 | 9,800 | 10,200 | 19,099 |
| PT Totals | 7,234 | 2,587 | 55,600 | 99,002 |
| VIT % | 27.43% | 51.31% | | |

---

[9] LDC = Left Dislocated Complement S_DIS = Dislocated Subject (postposed); S_TOP = Topicalized Subject (preposed); S_FOC = Focalized Subject (inverted); DisMods = Discontinuous Modifiers which include PP, PbyP, PofP, VP, RelCl, AP

| | PT % | 13.16% | 0.26% | | |
|---|---|---|---|---|---|

**Table 2.** Noncanonical structures and Unexpressed Subjects in VIT and PT

Total number of constituents for PT amounts to 720,086. Percent of NCS are computed on the number of Total Utterances, while percentage of Unexpressed Subjects are computed on the number of Total Simple Sentences. The nonprojectivity index for PT would then amount to 0.01004%. Expectancies for an Italian speaker for presence of a NCS are thus predictable to be fairly high, due to processing difficulties raised by number of Unexpressed Subjects(UnxSubj), in particular, as discussed above. This will not apply to an English speakers because NCS are unfrequent and used only in specific contexts and situations.

We will now take a few short examples of some of the most common noncanonical structures present in our treebank and will try to have them parsed by some of the best parsers of Italian[10]. As with previous examples, we show a syntactic augmented representation of the noncanonical structures which should be derived by the parsers in order to obtain a meaningful output. However, we limit ourselves to showing the most linguistically relevant cases, disregarding examples of discontinuities between head and predicative modifiers which are very frequent but do not add new information to the examples below. These are the examples we have chosen:

*1a. PP Adjunct Preposing in Participial Clauses*

Oggi ringrazio della cortesia in più occasioni dimostrata a me e ai miei colleghi./Today I thank for the courtesy in several cases shown to me and to my colleagues

This example is taken from bureaucratic documents where this structure is fairly common. The adjunct PP in più occasioni/in several cases is anteposed to the past participle dimostrata/shown from which it depends. The parser should thus be able to reject the processing of the PP inside the previous NP la cortesia/the courtesy, and also at the higher level where the governing finite verb ringrazio/thank is present, and leave it to the adjunct participle clause.

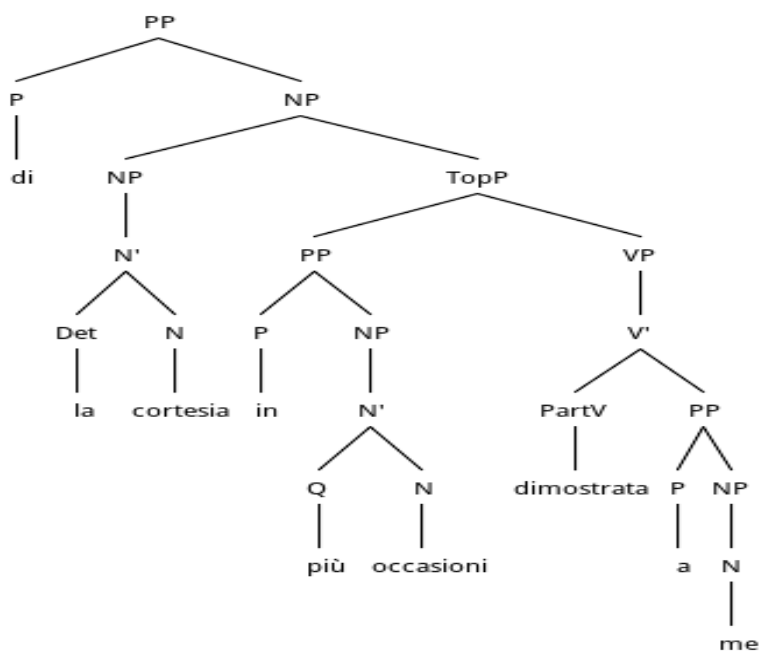

**Figure 14.** Tree structure for PP Adjunct Preposing in Participials

---

[10] http://tanl.di.unipi.it/it/ - DESR statistical only parser by Giuseppe Attardi. For a complete description of the parsing technique see https://sites.google.com/site/desrparser/
https://visl.sdu.dk/visl/it/parsing/automatic/parse.php Rule-based Constraint Grammar parser. Information on the parser can be found here, https://visl.sdu.dk/visl/about/ and the grammatical underlying theory here, https://visl.sdu.dk/visl2/constraint_grammar.html
http://tint.fbk.eu/ Italian version of Stanford Lexical Statistical Parser. The parser has been developed on the basis of Stanford Parser (see https://nlp.stanford.edu/software/lex-parser.shtml) by Giovanni Moretti and Alessio Palmero Aprosio. A complete description of the Italian version is available on the website.
http://textpro.fbk.eu/ modular parser by researchers at FBK in Trento. People contributing are quite a lot and can be found here, http://textpro.fbk.eu/people . A long list of publications relating the technical components of the parsers are to be found here, http://textpro.fbk.eu/publications . All the modules are statistically driven or make use of machine learning.
http://136.243.148.213/text_tree/ fully experimental parser by Matteo Grella that encompasses both a deep neural and a rule based approach. This parser is still in a prototype version.

*2a. Subject NP Extraposition*

Diventa così più acuta la contraddizione./(It) becomes thus more painful the contradiction

This example is another case of Subject Right Dislocation, and as such it is rejected by the call for canonical clauses. The main verb diventa /becomes is followed by a discourse marker, così/so and then in reverse order the arguments of the verb.

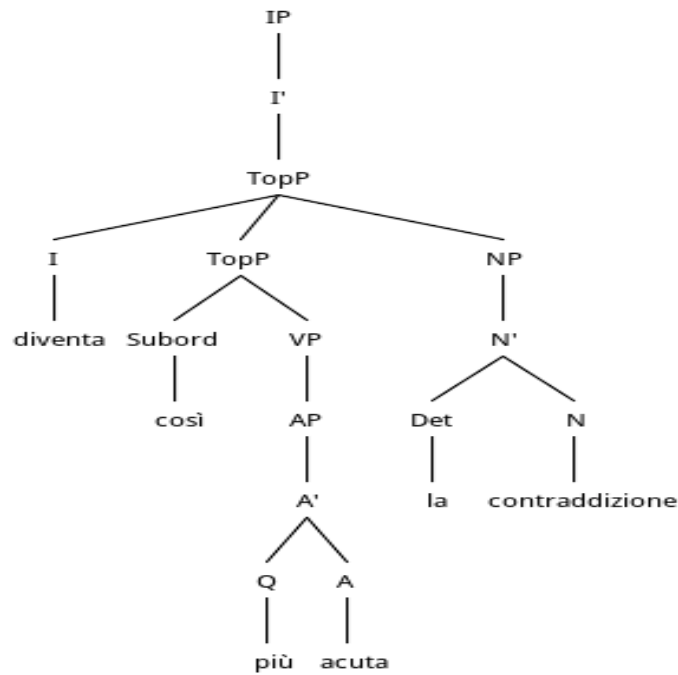

**Figure 15.** Tree structure for Subject NP Right Dislocation

*3a. Predicative AP Focussing in Fragments*

Buono invece in complesso il resto./Good on the contrary as a whole the rest.

This is a fragment where the implicit unexpressed verb can in fact be represented by verb to be, in its 3[rd] person singular, present indicative form, è/is. Fragments are computed after the clause has been fully parsed and no verb has been detected.

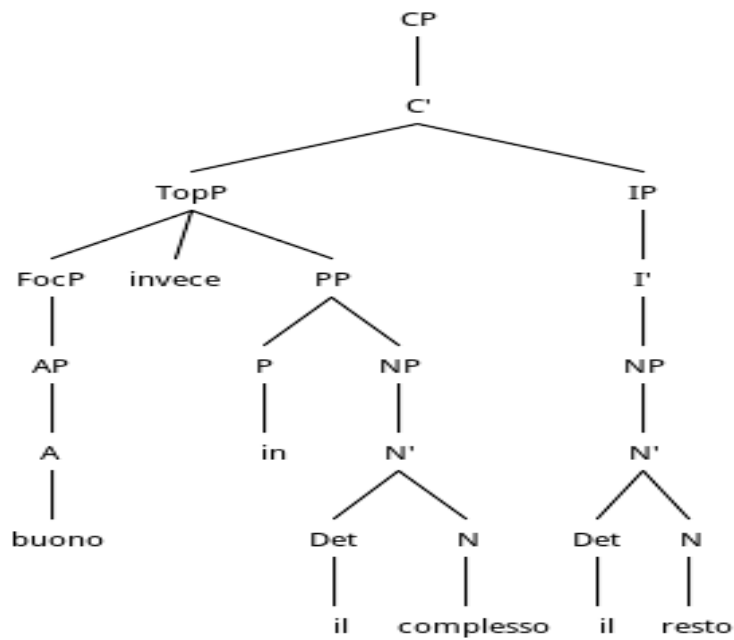

**Figure 16.** Tree structure for Predicative AP Preposing

*4a. Clitic Left Dislocation*

Una decisione importante Ghitti l'ha riservata a dopo le feste./A decision important Ghitti it has reserved to after the holidays.

This is a case of Clitic Left Dislocation where the Object NP decisione importante is fronted and has an indefinite determiner una. The NP is bound to an accusative clitic la thus transforming the NP from a FocP into a TopP. In fact the use of an abstract noun is by itself a case of topicalized assertion. From a computational point of view, we have again a non-canonical structure that requires the parser to fail an then to search the appropriate rule.

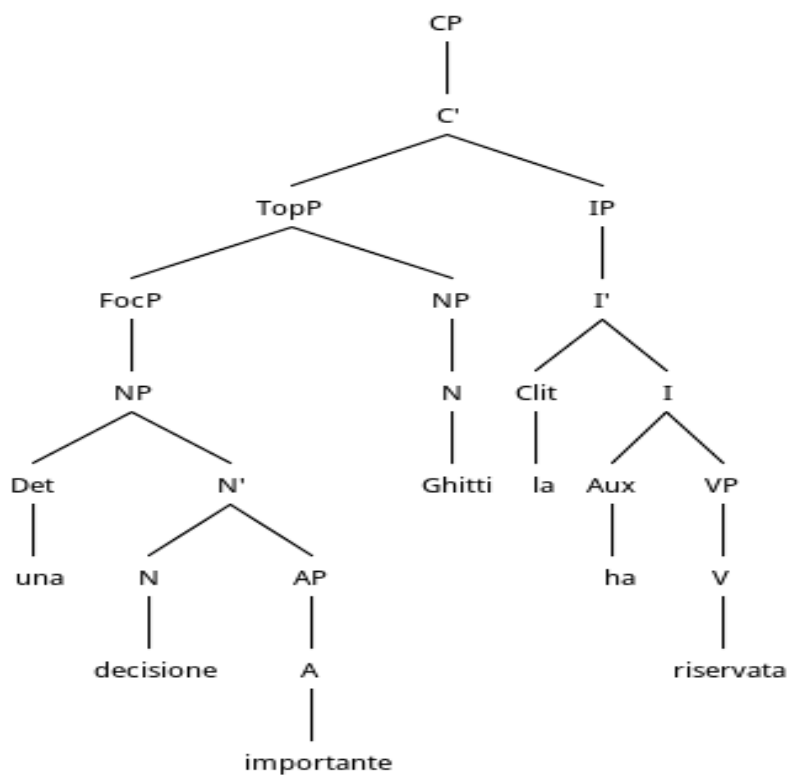

**Figure 17.** Tree structure for Clitic Left Dislocation

Here more examples which we don't represent in syntactic structures but which are similar to the ones already presented.

*5a. Predicative AP Focussing in Copulative Clauses*

L'importante ora è aprirlo di più./The important now is to open it more

*6a. Object NP Right Dislocation*

Le sue informazioni darebbero anche agli orientamenti di democrazia laica maggiori spinte./His information would give also to the orientations of secular democracy major thrusts

*7a. Parenthetical Insertion*

In questo libro Maria Teresa, spiegano alla Mondadori, darà esempi di carità concreti./In this book Maria Teresa, they explain at Mondadori, will give examples of charity concrete(plural)

Considering now example 7a., which is a case of parenthetical insertion before the IP structure and after the Subject NP Maria Teresa, it certainly constitutes a complex parsing example as the experiments below will show. The main difficulty for statistical parsers is to compute spiegano – 3[rd] person plural Present Indicative – as a subjectless verb due to lack of agreement with NP Maria Teresa. As will be shown below, no parser is able to compute Maria Teresa as Subject NP of darà/will give; and also no parser can compute spiegano as subjectless verb.

# 6. Parsing Complex Noncanonical Syntactic Structures

We ran our experiment using the same parsers both for poetic verses and nonpoetic noncanonical sentences. Here below a table that shows the results for poetic examples. We use abbreviated names for the parsers: TXP = TextPro from FBK Trento; VSL = the Danish Constraint Grammar based parser; TNT = the Tint parser always from FBK Trento but an Italian version of Stanford Lexicalized Parser; TLN = DeSR Attardi's parser from Pisa University.

| Sentence/Parser | TXP | VSL | TNT | TLN |
|---|---|---|---|---|
| *Sentence 1* | | | | |
| [FocP[NP Lei sola]] | *Subj | *X-!/0 | *Subj | *Comp |
| [Adv forse] | + | + | *AdvMod/sognatore | *Mod/sognatore |
| [Subj[NP il freddo | *Dobj | */ringrazio | */ringrazio | */ringrazio |
| sognatore]] | */e ai miei | */occasioni | */ringrazio | */ringrazio |
| [I educherebbe] | + | + | *ACL | + |
| [PP[P a][NP il tenero prodigio]] | + | + | + | + |
| *Sentence 2* | | | | |
| [C° ove] | + | + | + | + |
| [TopP[PP con te]] | + | + | + | */riprendere |
| [FocP[V° riprendere]] | + | *4/0 | + | */con |
| [I può] | + | *5/0 | *DOBJ | + |
| [VP a conversare] | + | *6/0-*7/5 | *ACL | + |
| [TopP[NP l' anima fanciulla]] | *Dobj *Subj-no | *9/7 | *Dobj *Subj-no | *Dobj *Subj-no |
| *Sentence 3* | | | | |
| Disse | + | + | + | + |
| [C° che] | + | + | + | + |
| [IP[I° gli hanno] | + | *4/0 | + | + |
| [TopP[NP il cor]] | + | *Subj | *Nsubj | */hanno |
| [PP di mezzo] | *mezzo-*Rmod/cor | *7/6 | *Nmod/cor | */cor |
| [NP[ il petto]] | *Subj | *Acc*10/4 | *Dobj | *Subj |
| [V°[PartV tolto]] | | | | |
| *Sentence 4* | | | | |
| [C° se] | + | + | + | + |
| [TopP[NP primavera]] | *VI-primavera | *ADVL | *Nmod/cuor | *Obj/Se |
| [TopP[NP il mio cuor | *Subj | *Subj | *Nmod/sordi | *Subj |
| [AP generoso]]] | + | + | + | + |
| [I° soffocasti] | + | + | *A | *RelCl/primavera |
| [PP di [NP spasimi] | + | + | *A | + |
| [AP sordi]]] | + | + | *S | + |
| *Sentence 5* | | | | |
| [C° né] | + | + | */dell'ora | + |
| [FocP[NP l' oblïoso incanto]] | *Subj | + | *Subj | *Subj |
| [TopP[PP [P di] [NP l' ora]]] | *Rmod-incanto | *Subj | *V | + |
| [NP il ferreo bàttito] | + | + | *Dobj | *Conj |
| [I° concede] | + | + | *Advcl | + |

**Table 3.** Comparative Analysis of the output of rhetorical figures texts by four parser of Italian

We decomposed the text of the five longest and sentence-like verses into 31 constituents and assigned a + sign to those that received a correct analysis and a starred comment to the rest. If needed, the comment is split into two by a slash to indicate the wrong head the constituent has been made dependent of. In some cases, both dependency and grammatical relation are wrong. In some other cases both tag and relation are wrong. They will be counted double. Eventually, I marked as mistake the lack of a Subject in the sentence. As a general remark, we note that in dependency structure theory only lexically expressed items can be part of the output. As a result, none of the systems introduce empty categories in their representation apart from VSL which however does not make any use of it in these examples. This goes against the need for a Subject to be always present as a general principle of any linguistic theory. In particular then, whenever it is a Topic and more so when it is a non-overt unexpressed continuing Topic from a previous utterance in the discourse.

| Parser | No.Errs | %correct |
|---|---|---|
| TXT | 15 | 51.61 |
| VSL | 18 | 41.94 |
| TNT | 22 | 29.03 |
| TLN | 15 | 51.61 |

**Table 4.** Performance of four parsers of Italian on noncanonical verse

As can be gathered from Table 4. the level of performance is slightly higher than 50% (48.39% errors for either wrong label, or wrong dependency) for two parsers, TextPro and TALN. However, none of the parsers has been able to compute noncanonical elements correctly, be it a single word or a constituent.

What we will do now is using the same parsers for parsing noncanonical structures in written Italian as they appear in newspaper texts. The results of the analysis are reported here below in Table 5 where we listed 5 sentences and isolated 29 constituents. Results are better than before but again only slightly over 50% correct structures.

| Sentence/Parser | TXP | VSL | TNT | TLN |
|---|---|---|---|---|
| *Sentence 1* | | | | |
| Ringrazio | + | + | + | + |
| della cortesia | *Rmod | *Subj | *Nmod | + |
| in più occasioni | *Dobj | */ringrazio | */ringrazio | */ringrazio |
| dimostrata | */e ai miei | */occasioni | */ringrazio | */ringrazio |
| a me e | + | */0 | + | + |
| ai miei colleghi | *NotConn | + | + | + |
| *Sentence 2* | | | | |
| diventa | + | + | + | + |
| così | + | */più | */acuta | + |
| più | */così | + | + | + |
| acuta | + | *ADJ | + | + |
| la contraddizione | + | *NPHR-noSUBJ | *DOBJ | + |
| *Sentence 3* | | | | |
| una decisione | *Subj | *Subj | *NSubj | *Root |
| importante | */Ghitti | + | + | */Ghitti |
| Ghitti | */decisione | + | */decisione | */decisione |
| l'ha riservata | + | + | + | *RELCL |
| a dopo le feste | + | + | + | *a |
| *Sentence 4* | | | | |
| Le sue informazioni | + | + | + | + |
| darebbero | + | + | + | + |

| anche agli orientamenti | + | + | + | + |
| di democrazia laica | + | + | + | + |
| maggiori | */orientamenti | + | */democrazia | + |
| spinte | *Rmod*/darebbero | + | */democrazia | */democrazia |
| *Sentence 5* | | | | |
| in questo libro | */spiegano | */little_pro | */spiegano | */spiegano |
| Maria Teresa, | *Appos*/libro | *PROP*/libro | Subj-*/spiegano | *Appos*/libro |
| spiegano | *Root | *Root | *Root | *Root |
| alla Mondadori, | + | + | + | + |
| darà | *DOBJ*/spiegano | + | */spiegano | */spiegano |
| esempi di carità | + | + | + | + |
| concreti | */carità | *0 | */carità | */carità |

**Table 5.** Comparative Analysis of the output of noncanonical written texts by four parsers of Italian

Results of the analysis show that in this case, parsers have been able to identify at least one of the noncanonical structures, but not the same one. In particular, both TXT and TLN have correctly labeled as SUBJ la contraddizione and as Predicative Adjective più acuta in sentence 2. Then VSL has correctly identified Ghitti as the SUBJ in sentence 3 but it has done so by inserting another empty subject at the beginning of the sentence and labeling una decisione also as Subject – whereas only one should have been identified as such. However, VSL is the only parser that has been able to label correctly sentence 4. Here below the overall evaluation

| Parser | No.Errs | %correct |
|---|---|---|
| TXT | 18 | 37.03 |
| VSL | 14 | 51.73 |
| TNT | 14 | 51.73 |
| TLN | 14 | 51.73 |

**Table 6.** Performance of four parsers of Italian on noncanonical written text

In what follows we will be using the system *GETARUN*, which has been created in the LLC – Laboratory of Computational Linguistics – of Ca Foscari University, Venice. The system is organized in a pipeline which requires the completion of certain levels of representation in order to continue the analysis which goes from tokenization to syntax, semantics and, finally, the discourse model. In particular, after tokenization and sentence splitting, each sentence is passed to the first level of analysis where tagging is performed. This is followed by a syntactic parser that works depth-first deterministically, using a variable length lookahead mechanism to prevent producing hypothesis which are not attested in the string. The idea is that of an incremental parser, trying to mimick as faithfully as possible the analysis steps performed by a human parser while reading/hearing the sentence. Words are parsed from left to right using psychologically relevant information like what are the most frequent syntactic structures given a certain sequence of words. As an intuitive probability estimation, the parser preferences are set to parse canonical structures at first, up to an unwanted or unexpected constituency fault. This will happen every time the current structure does not allow continuation and interpretation attempts have to be discarded causing a local failure. Following LFG but also Minimalist Theory(hence MT), in order for constituents to be passed from Short to Long Memory, they need to be interpreted: this corresponds to f-structures in LFG and to a Phase in MT(see [10]:chapt.3) which is constituted by a fully realized argument structure preceded by a Verb. Failures will correspond to the longer lag ensuing from presence of a noncanonical structure, be it a simple discontinuity or a displaced constituent to the right or to the left of the sentence. This is what the system should be able to show in computing time, an increase which should make a notable difference from a canonical structure that has an identical number of words or heads. To verify our hypothesis we have run *GETARUN* on the previous examples and taken note of computing time at every step of the computation. As said above, the system is organized into a pipeline where the first module computes c-structure and f-structure before delivering the output to the level of semantic processing. *GETARUN* parses the whole sequence of 12 sentences in 2 seconds producing a complete semantic and pragmatic analysis. Sentences requiring the longest computing time are sentence n. iii that we also analyse below, and sentence n. 7 above. Number of errors made are 4 as a whole, i.e. considering both syntax and semantics.

We report here below results only for some of the examples, at first the f-structure computed for sentence n. xii (Lei sola forse il freddo sognatore educherebbe al tenero prodigio).

```
indice:f15
pred:educare
lex_form:[np/subj1/causer_emot/[+hum],np/obj1/theme_emot/[-ani,+hum],pp/obl/exper/a/[]]
voice:attiva
modo:cond
tempo:pres
sogg/theme_emot:indice:sn13
        pred:sognatore
        gen:mas
        num:sing
        pers:3
        spec:def:+
        mods:mod_a:indice:sa11
                pred:freddo
        tab_ref:[+ref,-pro,-ana,+class]
focus/theme_emot:indice:sn12
         cat:[umano]
         pred:lei
         pers:3
         gen:fem
         num:sing
         spec:def:+
         mods:mod_a:indice:sa9
                cat:[estens]
                pred:solo
                gen:fem
                num:sing
           mod_a:indice:sa9
                cat:[estens]
                pred:solo
                gen:fem
                num:sing
         tab_ref:[+ref,+pro,-ana,+me]
         antecedente:esterno
         interpretazione:specifica
adjs:adj/modal:pred:forse
     adj/exper:sem_mark:a
         ogg/nil:indice:sn14
                pred:prodigio
                gen:mas
                num:sing
                pers:3
                spec:def:+
                mods:mod_a:indice:sa12
                    cat:[valutativo]
                    pred:tenero
                    gen:mas
                    num:sing
                tab_ref:[+ref,-pro,-ana,+class]
```

**Figure 18.** F-structure representation of the sentence "Lei sola forse il freddo sognatore educherebbe al tenero prodigio"

Elapsed time CIC1: **0.07065**
Elapsed time CIC2: **0.00209**
Elapsed time CIC3: **0.00582**

We are using three computing complexity indices which are situated inside the pipeline and are able to record start and end computing time of each module. The computation is measured in msecs. CIC1 records computing time required for all parsing processes including tagging, syntactic c-structure building and f-structure mapping; CIC2 registers the following module, where quantifiers raising and pronominal binding are performed. CIC3 eventually computes the final higher discourse modules where anaphora and coreference resolution is done, logical form is created and spatiotemporal locations are asserted and inferred if needed. The final computation regards the building of the model in line with Situation Semantics which transforms the semantic output into a list of facts containing an information unit called infon each. Infons can be properties, relations, individuals or classes each one associated to a unique index, a semantic identifier in the discourse model. Each fact is also accompanied by a polarity and a spatiotemporal couple of indices. Arguments of predicates are associated to a semantic role or a semantic class. Eventually discourse structure is computed at clause level.

**ind(infon1,id1)**
**fact(infon2,solo,[ind:id1],1,univ,univ)**
**fact(infon3,isa,[ind:id1,class:indefinite],1,univ,univ)**
**fact(infon4,inst_of,[ind:id1,class:donna],1,univ,univ)**
**ind(infon5,id2)**
**fact(infon6,freddo,[ind:id2],1,univ,univ)**
**fact(infon7,inst_of,[ind:id2,class:astratto],1,univ,univ)**
**fact(infon8,isa,[ind:id2,class:sognatore],1,univ,univ)**
**ind(infon9,id3)**
**fact(infon10,tenero,[ind:id3],1,univ,univ)**
**fact(infon12,inst_of,[ind:id3,class:astratto],1,univ,univ)**
**fact(infon13,isa,[ind:id3,class:prodigio],1,univ,univ)**
**fact(id4,educare,[causer_emot:id2,theme_emot:id1],1,tes(f15_sent1),univ)**
**fact(infon16,isa,[arg:id4,arg:st],1,tes(f15_sent1),univ)**
**fact(infon17,isa,[arg:id5,arg:tloc],1,tes(f15_sent1),univ)**
**fact(infon18,pres,[arg:id5],1,tes(f15_sent1),univ)**
**fact(infon19,time,[arg:id4,arg:id5],1,tes(f15_sent1),univ)**
**fact(infon20,forse,[arg:id4],1,tes(f15_sent1),univ)**
**fact(infon21,focus,[arg:id1,arg:id4],1,tes(f15_sent1),univ)**
**fact(infon23,a,[arg:id4,nil:id3],1,tes(f15_sent1),univ)**

**Figure 19.** Model in Situation Semantics style for sentence n. xii

The additional information we have now is the predicate-argument structure associated to educare and the properties associated to sets and individuals: for instance, we know that the focussed item lei is an undefined woman. Besides, the focussed item has as its argument id4, the index assigned to the main predicate, educare, i.e. the main relation. Every infon has a polarity, here the value is 1, i.e. positive; and the final two indices are the temporal and the spatial ones.

We will now show the output of another example, n. x., where we have both an extraposed Subject and a preposed Object, and where we don't expect longer computing times.

```
indice:f10
main/prop:[]
adj:sem_mark:se
   sub/prop:indice:f9
      pred:soffocare
      lex_form:[np/subj1/agent/[+hum],np/obj1/theme_aff/[-ani,+hum,+abst]]
      voice:attiva
      modo:ind
      tempo:pres
      focus/agent:indice:sn13
         pred:cuor
         gen:mas
         num:sing
         pers:3
         spec:def:+
         sogg/poss:indice:sn14
            pred:mio
            gen:mas
            num:sing
            spec:def:+
            tab_ref:[+ref,+pro,-ana,+me]
            antecedente:esterno
            interpretazione:definita
         mods:mod_a:indice:sa16
             pred:generoso
             gen:mas
             num:sing
         tab_ref:[+ref,-pro,-ana,+class]
      topic/theme_aff:indice:sn12
          pred:primavera
          pers:3
          spec:def:'0'
          tab_ref:[+ref,-pro,-ana,+class]
      adjs:adj/matter:sem_mark:di
          ogg/matter:indice:sn15
             pred:spasimo
             gen:mas
             num:plur
             pers:3
             spec:def:'0'
             mods:mod_a:indice:sa17
                 cat:[valutativo]
                 pred:sordo
                 gen:mas
                 num:plur
             tab_ref:[+ref,-pro,-ana,+class]
```

**Figure 20.** F-structure representation for sentence "Se primavera il mio cuor generoso soffocasti di spasimi sordi"
Elapsed time CIC1: **0.05961**
Elapsed time CIC2: **0.00209**
Elapsed time CIC3: **0.07280**

Eventually we show the output of sentence no. Iii. (Penso a un verde giardino ove con te riprendere può a conversare l'anima fanciulla.) which contains a relative clause where however the pronominal head is unambiguously interpreted as a locative pronoun. To allow for morphological features to appear on the same page we have compacted them on the same line.

```
indice:f2
pred:pensare
lex_form:[sn/sogg/actor/[umano,animato],sp/obl/tema/a/[luogo,oggetto,parte,place,solido]]
voice:active
modo:ind,tempo:pres
cat:ment_perloc
sogg/actor:indice:sn2
     cat:[umano,animato]
     pred:pro
     pers:1,spec:def:+,caso:nom
     tab_ref:[+ref,+pro,-ana,-me]
     antecedente:esterno,interpretazione:definita
obl/tema:indice:sn3
     cat:[luogo,oggetto,parte,place,solido]
     pred:giardino
     gen:mas,num:sing,pers:3,spec:def:-,tab_ref:[+ref,-pro,-ana,+class]
     qmark:q1
     mods:relativa:topic:tipo_topic:relativo
             indice:sp1
             cat:[luogo,oggetto,parte,place,solido]
             pred:ove
             pers:3,gen:mas,num:sing,caso:obl, tab_ref:[+ref,-pro,-ana,-me]
             controllore:sn3
         indice:f1
         pred:riprendere
         lex_form: [np/subj1/theme_unaff/[+hum],vinf/vcomp/prop/a/[subj=subj1]]
         voice:active
         modo:ind,tempo:pass_pross
         supporto:potere
         cat:risultato
         sogg/agente:indice:sn12
               cat:[animato,istituzione,oggetto,parte,umano]
               pred:anima
               gen:fem,num:sing,pers:3,spec:def:+,tab_ref:[+ref,-pro,-ana,+class]
               adjs:ncomp/tema:indice:sn13
                     cat:[animato,istituzione,oggetto,umano]
                     pred:fanciulla
                     gen:fem,num:sing,pers:3,spec:def:'0'
                     tab_ref:[+ref,-pro,-ana,+class]
         vcomp/prop:indice:finf1
               pred:conversare
               lex_form:[np/subj1/actor/[+hum]]
               modo:inf,tempo:pres
               cat:activ
               sogg/creator:indice:sn11
                     cat:[]
                     pred:pPro
                     tab_ref:[+ref,+pro,+ana,-me]
                     antecedente:sn4,interpretazione:definita
               adjs:adj/locativo:sem_mark:in
                     ogg/nil:indice:sn6
                           cat:[luogo]
                           pred:vbl
                           controllore:sp1
                           tab_ref:[+ref,-pro,-ana,-me]
         adjs:adj/circumst:sem_mark:con
               indice:sn4
               cat:[umano]
               pred:te
               pers:2,gen:any,num:sing,spec:def:+,caso:obl,tab_ref:[+ref,+pro,-ana,+me]
               antecedente:esterno,interpretazione:definita
     mod_a:indice:sa2
         cat:[]
         pred:verde
```

**Figure 21.** F-structure representation for sentence "Penso a un verde giardino ove con te riprendere può a conversare l'anima fanciulla"

Elapsed time CIC1: **0.05486**
Elapsed time CIC2: **0.01068**
Elapsed time CIC3: **0.01260**

As can be noticed, all computing times have increased remarkably, apart from the parser. It is important to distinguish the parsing process required for this type of structures, where in addition to the presence of a relative clause, we have a displaced constituent, i.e. a right dislocated NP which has to be interpreted as SUBJect of its internal verb predicate. At first, the system fails to produce a sensible f-structure interpretation in which canonical transitive sentences are organized in a sequence. The intervening infinitival clause contributes to complicate the interpretation process due to the need to assign a controller to the SUBJect bigPRO pronoun. As can be seen, the bigPro subject of the infinitival, indexed sn11, is wrongly controlled by sn4 – the pronoun te -, and not by l'anima fanciulla, indexed sn12.

Here below a table with the complete computing times for most of the structures we discussed above, where we used the following abbreviations: S=subject; O=object; iO=indirectObject; Obl=oblique; V=verb; Adj=adjunct; Mo=modifier; Cl=clitic; Ax=auxiliary; Cg=subordinating conjunction; Pa=parenthetical

| Poems/Cics | No. Toks | Basic config. | CIC1 | CIC2 | CIC3 | TOTAL |
|---|---|---|---|---|---|---|
| 2a.Contraddizione | 6 | VCgPaS | 0.01156 | 0.00099 | 0.00537 | 0.01792 |
| 3. Anima fanciulla | 15 | OblVVS | 0.05486 | 0.01068 | 0.01260 | 0.07814 |
| 4. Di mezzo il petto | 11 | ClAxOOblV | 0.09071 | 0.00439 | 0.04819 | 0.14329 |
| 4a. Ghitti | 11 | OSClVAdj | 0.03431 | 0.00167 | 0.01609 | 0.05207 |
| 6a. Orientamenti | 13 | SViOO | 0.02015 | 0.00327 | 0.03388 | 0.05730 |
| 7a. Maria Teresa | 14 | AdjSPaVO | 0.06702 | 0.00411 | 0.02402 | 0.09515 |
| 10. Cuore soffoca | 10 | SOViO | 0.05961 | 0.00209 | 0.07280 | 0.13450 |
| 11. Ferreo battito | 11 | OMoSV | 0.09733 | 0.00120 | 0.00959 | 0.10812 |
| 12. Lei sola | 10 | OSViO | 0.07065 | 0.00209 | 0.00582 | 0.07856 |

**Figure 22.** Computing times for 9 sentences and their basic configurations

Longest computing time are associated to noncanonical sentences with two important features: they are only found in poetry and when the discontinuity takes place at the beginning of the sentence (CP) and not in the complement section (VP). Subject inversion does not induce any delay in computation as can be seen from 2a. Example 3 – the example with the highest number of tokens - is another similar case, with Subject NP is in VP internal position. Also example 4a, a case of Clitic Left Dislocation which requires failure at the level of canonical structures, but its computing time is identical to 6a a canonical sentence which only has a scrambled object NP. The worst case is constituted by example 4, which has an auxiliary separated from its main past participle verb. This is followed by example 10, a Subject+Object type of non-canonical sentence. Example 12 is also a case of fronted NPs and CIC1 testifies the increase in computing time, but then the semantics is fairly straightforward and the overall computation is almost identical to example 3, the only example that has a relative clause.

# 7. Conclusion

In this paper we have been presenting and discussing in detail the peculiar linguistic nature of some of the most interesting rhetorical figures present in poetry written by Italian poets beginning and middle of '900. We have been able to trace some of the syntactic structures back to previous Italian poets of the past, but in some cases of hyperbaton we had to trace back its origins to Latin verse or text. This is no surprise seen that Italian is a direct descendant of that language. The way in which some hyperbaton has been used in Italian is however not usually attested in Latin – where the auxiliary is not raised but is left at the end of the sentence. We have privileged the introduction of LFG as a linguistic theory able to explain the psychological relevance of noncanonical structures and its impact in informational terms. The representations we used were mostly constituency and syntactic projections in terms of X-bar theory and some functional representations in which only linguistic features and grammatical relations are present. Eventually, we presented data from our treebank – VIT (Venice Italian Treebank) – to illustrate the use of noncanonical structures in written text from Italian newspapers. We compared their frequency to that present in a famous American English treebank, the Penn Treebank. The final two sections are purely computational: the first one is an experiment in which we used the best four online parsers of Italian to parse the sentence-like excerpts from poetry and the ones we collected from our treebank of written Italian newspaper texts. The experiment showed how statistical-probabilistic parsers using models taken from standard Italian are unable to cope with noncanonical structures, both the exceptional ones from poetry, and the ones commonly used in standard Italian. The second experimental section uses *GETARUN*, the symbolic and rule-based system of Italian which has been created at the LLC of Ca Foscari. Here we shows how the parsing of noncanonical structures may involve a greater computing time. This is motivated by psycholinguistic and processing issues which are thus both confirmed also by the use of LFG. In particular, commonly used noncanonical structures show a much lower increase in computational time than those found in poetry. Finally, only when noncanonical structures are constituted by fronted linguistic material computation time increases remarkably.